\def\paperTitle{Modern Self-Supervised Vision Transformers Are Pareidolic}

\def\authorBlock{
{\fontsize{12}{16}\selectfont
    Zihang Lin\textsuperscript{1,2,*}%
\qquad
    Huaiyuan Qin\textsuperscript{1,$\dagger$}%
\qquad
    Liangli Zhen\textsuperscript{1}%
\qquad
    Qing Guo\textsuperscript{3}%
}\\
{\fontsize{12}{16}\selectfont
    Jun Liu\textsuperscript{4}%
\qquad
    Yew Soon Ong\textsuperscript{1,2}%
\qquad
    Hongyuan Zhu\textsuperscript{1,\Letter}%
}\vspace{2.5pt}\\ 

{\fontsize{10.8}{16}\selectfont
\textsuperscript{1}Institute of Advanced Intelligence and Computing (IAIC), A*STAR, Singapore
}\\

{\fontsize{10.8}{16}\selectfont\textsuperscript{2}Nanyang Technological University 
}\hspace{9pt}
{\fontsize{10.8}{16}\selectfont\textsuperscript{3}Nankai University}
\hspace{9pt}
{\fontsize{10.8}{16}\selectfont\textsuperscript{4}Lancaster University
}\\

{\tt\fontsize{9.6}{16}\selectfont
linz0071@e.ntu.edu.sg, \{qinhy, zhuh\}@a-star.edu.sg\hspace{-6.28pt}}\\
}

\pdfoutput=1
\documentclass[10pt,twocolumn,letterpaper]{article}
\usepackage{meta/paper}

\usepackage{fix-cm}
\usepackage{array}
\usepackage{bbm}

\usepackage{nicematrix}

\usepackage{tipa}
\usepackage{dsfont}
\usepackage{etoolbox}

\usepackage[noend]{algorithmic}
\usepackage{algorithm}

\usepackage{float}
\usepackage{newfloat}
\usepackage{listings}
\floatstyle{ruled}
\newfloat{listing}{tb}{lst}{}
\floatname{listing}{\small Algorithm}

\definecolor{mygray}{RGB}{234,234,234}

\usepackage{adjustbox}
\usepackage[table,dvipsnames]{xcolor}

\definecolor{darkgreen}{rgb}{0.13, 0.55, 0.13}

\usepackage{graphicx}	
\usepackage{amsmath}	
\usepackage{amssymb}	
\usepackage{booktabs}
\usepackage{times}
\usepackage{epsfig}
\usepackage{caption}
\usepackage{float}
\usepackage{placeins}
\usepackage{color, colortbl}
\usepackage{stfloats}
\usepackage{enumitem}
\usepackage{tabularx}
\usepackage{xstring}
\usepackage{multirow}
\usepackage{xspace}
\usepackage{url}
\usepackage{subcaption}
\usepackage{xcolor}

\definecolor{Gray}{gray}{0.5}
\definecolor{nicergreen}{rgb}{0.13, 0.54, 0.13}
\definecolor{nicered}{rgb}{0.83, 0.16, 0.16}
\definecolor{lightgray}{RGB}{230, 230, 230}
\definecolor{Highlight}{HTML}{39b54a}  %

\usepackage{appendix}
\usepackage{footnote}

\usepackage{microtype}
\usepackage{cuted}
\usepackage{tocloft}

\usepackage[T1]{fontenc}
\usepackage{DejaVuSans}

\usepackage{ifsym, marvosym}

\let\svthefootnote\thefootnote
\newcommand\freefootnote[1]{%
  \let\thefootnote\relax%
  \footnotetext{#1}%
  \let\thefootnote\svthefootnote%
}

\newcommand{\ours}{PSC\xspace}
\newcommand{\oursfull}{Pareidolic Subspace Calibration\xspace}
\newcommand{\oursfullbold}{\textbf{P}areidolic \textbf{S}ubspace \textbf{C}alibration\xspace}

\definecolor{cDINO}{HTML}{6E86A6}
\definecolor{cDINOtwo}{HTML}{C99A6B}
\definecolor{cDINOthree}{HTML}{7FA394}
\newcommand{\gain}[1]{\rlap{\textsubscript{\fontsize{6pt}{1em}\selectfont\textcolor{nicergreen}{$+$\textbf{#1}}}}}

\newcommand{\gray}[1]{\textcolor{gray}{#1}}

\newcommand{\cls}{\texttt{[CLS]}\xspace}
\newcommand{\Vs}{V_s}
\usepackage{tikz}
\usetikzlibrary{calc}

\definecolor{linkblue}{rgb}{0.21,0.49,0.74}
\usepackage[pagebackref,breaklinks,colorlinks,allcolors=linkblue]{hyperref}
\usepackage[capitalize]{cleveref}
\crefname{section}{Sec.}{Secs.}
\crefname{table}{Table}{Tables}
\crefname{figure}{Fig.}{Figs.}

\begin{document}
\title{\paperTitle}
\author{\authorBlock}
\maketitle
\addtocontents{toc}{\protect\setcounter{tocdepth}{-1}}

\freefootnote{\hspace{-12pt}\textsuperscript{*}This paper was completely accomplished when ZL interned at A*STAR.}
\freefootnote{\hspace{-12pt}\textsuperscript{$\dagger$}Project lead. }
\freefootnote{\raggedright\settowidth{\dimen0}{\textsuperscript{\Letter}}\hangindent=\dimexpr1.8em-12pt+\dimen0\relax\hangafter=1\hspace{-12pt}\textsuperscript{\Letter}Corresponding author, with responsibility for conceptualization and scientific direction.}
\makeatother

\begin{strip}
  \centering
  \vspace{-42pt}
  \includegraphics[width=\textwidth]{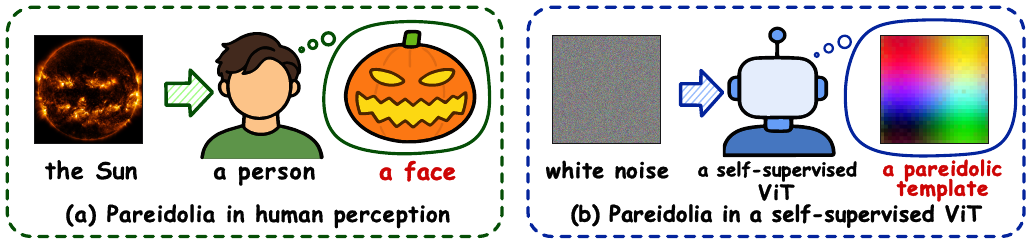}
  \vspace{-18pt}
  \captionof{figure}{\textbf{Modern self-supervised ViTs are pareidolic.} (a) People
  perceive a meaningful pattern in a random or ambiguous stimulus, such as a
  jack-o'-lantern face in an image of the Sun, whose active regions happen to form
  eyes and a mouth. (b) A self-supervised ViT imposes a spatially coherent pattern
  on white noise: the features of DINOv3-B on one white-noise image along their three
  leading directions, shown as red, green and blue. We call such a template
  pareidolic.}
  \label{fig:teaser}
  \vspace{-6pt}
\end{strip}

\begin{abstract}
In human perception, pareidolia is the tendency to perceive a meaningful pattern,
such as a face, in a random or ambiguous stimulus.
We find a similar phenomenon in modern self-supervised Vision Transformers (ViTs):
on white noise, which contains neither objects nor spatial structure, a few
directions of their features still form spatially coherent patterns across the
patches, imposed by the features themselves.
We call this behavior pareidolia, and the span of these directions the pareidolic
subspace.
On natural images, it holds most of the feature variance in DINOv2 and DINOv3 but
little in DINO.
This excess explains why DINOv2 and DINOv3, despite stronger dense features, fall
far behind DINO in object emergence, the hallmark ability of self-supervised ViTs
to identify the patches of an object on their own.
Our experiments further show that scaling down their pareidolic subspaces steadily
restores their ability to discover objects, so their object emergence is hindered
rather than lost.
These findings naturally lead to a training-free calibration: \oursfullbold
(\ours) measures the pareidolic subspace only on a noise input and then calibrates
the features of natural inputs by scaling its directions down to the variance level
of the remaining directions.
Across diverse object emergence tasks, \ours improves DINOv2 and DINOv3 by up to
31.7 CorLoc on object discovery, 30.2 maxF on saliency detection and
59.9~$\mathcal{J}$ on video object segmentation.
To identify the cause of pareidolia, we trace it block by block and find that it
forms in the middle blocks of every self-supervised model we examine but stays
strong to the last block only in models trained with a patch-level loss;
controlled training confirms that an image-level loss alone lets it fade.
\end{abstract}

\section{Introduction}
\label{sec:intro}

\begin{figure*}[t]
  \centering
  \includegraphics[width=\textwidth]{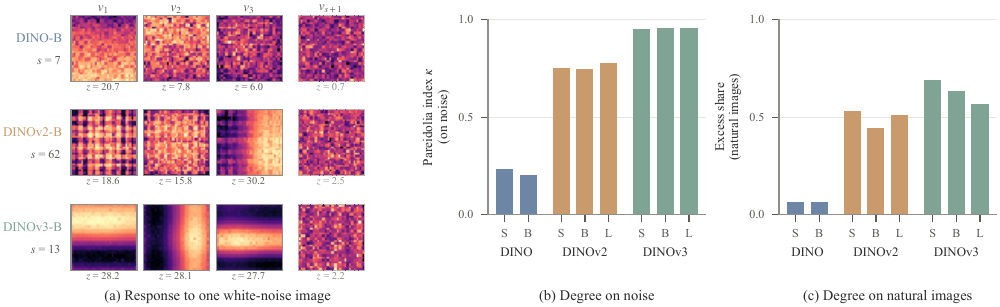}
  \vspace{-18pt}
  \caption{\textbf{On white noise, the features impose a template, faint in DINO and
  strong in DINOv2 and DINOv3.} (a) The response of the Base models to one
  white-noise image along their first three directions (\cref{eq:svd}) and along the
  first direction after the pareidolic subspace (dashed frame), with Moran's $z$ under
  each map (\cref{eq:moranz}) and the number $s$ of directions in the subspace.
  (b) The pareidolia index $\kappa$ on noise (\cref{eq:kappa}) and (c) the excess
  share of the pareidolic subspace on VOC07 images, for all eight models.}
  \vspace{-13pt}
  \label{fig:probe}
\end{figure*}

Pareidolia is the tendency to perceive a meaningful pattern, such as a face, in a
random or ambiguous stimulus~\cite{zhou2020you}, like the jack-o'-lantern face that
people perceive in an image of the Sun (\cref{fig:teaser}a).
In human perception, pareidolia arises when learned expectations dominate weak or
ambiguous input~\cite{liu2014seeing,lhotka2023no}, and it is studied with pure
noise, where the input offers no evidence at all: observers shown pure noise report
faces and letters~\cite{liu2014seeing}, and averaging the noise images that
triggered their reports recovers the internal representation behind them, which
psychophysics models as a template that the observer matches against the
input~\cite{gosselin2003superstitious,murray2011classification}.

We run the same test on modern self-supervised Vision Transformers
(ViTs)~\cite{dosovitskiy2021image} and find a similar phenomenon
(\cref{fig:teaser}b).
White noise contains neither objects nor spatial structure, so the features it
produces should have no spatial structure either.
Yet the leading directions of their features form spatially coherent patterns
across the patches, such as smooth waves across the image (\cref{fig:probe}a): a
template that the features impose, as human observers impose theirs on noise.
We therefore call this behavior pareidolia, and the span of these directions the
pareidolic subspace.
This pareidolia varies in degree: it is faint in DINO~\cite{caron2021emerging} and
strong in its successors, DINOv2~\cite{oquab2024dinov2} and
DINOv3~\cite{simeoni2025dinov3} (\cref{fig:probe}b).
We study the DINO lineage, in which object emergence was first observed, and find
pareidolia in other self-supervised ViTs as well (supplementary
\cref{sec:supp-kappa}).
On natural images, the pareidolic subspace holds most of the feature variance in
DINOv2 and DINOv3 but little in DINO: for the Base models, 8\% of the directions
hold 53\% of the variance in DINOv2, and 2\% hold 65\% in DINOv3
(\cref{fig:probe}c).

\begin{figure}[t]
  \centering
  \includegraphics[width=0.9\linewidth]{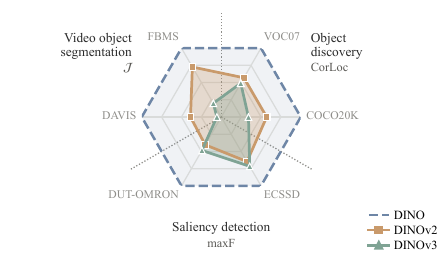}
  \vspace{-9pt}
  \caption{\textbf{Object emergence fades in DINOv2 and DINOv3.} Object emergence of
  \textcolor{cDINOtwo}{DINOv2} and \textcolor{cDINOthree}{DINOv3} relative to
  \textcolor{cDINO}{DINO (outer hexagon)}, as family means over model sizes. Each
  task takes one sector, labelled outside with its metric, and two datasets,
  labelled at the vertices.}
  \label{fig:radar}
  \vspace{0.3em}
  \includegraphics[width=0.9\linewidth]{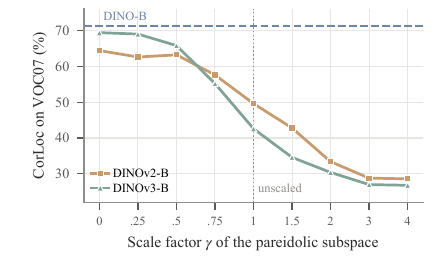}
  \vspace{-9pt}
  \caption{\textbf{Scaling down the pareidolic subspace restores object discovery.}
  Object discovery on VOC07 images with the coordinates of DINOv2-B and DINOv3-B
  along their own pareidolic subspaces scaled by a factor $\gamma$; $\gamma=1$ leaves the
  features unchanged, and the dashed line is DINO-B.}
  \vspace{-21pt}
  \label{fig:scaling}
\end{figure}

This excess matters for object emergence, the hallmark ability of self-supervised
ViTs to identify the patches of an object on their own~\cite{caron2021emerging}.
Object emergence is revealed by grouping patches according to the cosine
similarity of their final-layer keys, the very features in which we find
pareidolia, and it is tested by three standard tasks: unsupervised object
discovery~\cite{simeoni2021localizing,wang2022self,wang2023cut}, saliency
detection~\cite{wang2022self,simeoni2023unsupervised} and video object
segmentation~\cite{wang2023tokencut}.
When the pareidolic subspace holds most of the variance, it dominates these
similarities, so patches are grouped by the pareidolic pattern rather than by the
objects they belong to.
Accordingly, DINOv2 and DINOv3, despite stronger dense features and large gains on
dense tasks solved by trained heads, fall far behind DINO in object emergence
across diverse tasks (\cref{fig:radar}).

To examine how the pareidolic subspace affects object emergence, we rescale it in
DINOv2 and DINOv3: scaling it down steadily restores their object discovery, and
scaling it up degrades it further (\cref{fig:scaling}).
Object emergence in DINOv2 and DINOv3 is therefore hindered by pareidolia rather
than lost.
These findings naturally lead to a training-free calibration: since the pareidolic
subspace is a property of the model and appears without any image content, its
directions and their excess variance can be estimated from the
response to noise alone.
We propose \oursfullbold (\ours), which measures the pareidolic subspace only on a
noise input and then calibrates the features of natural inputs by scaling its
directions down to the variance level of the remaining directions, without training
or labels.
Across diverse object emergence tasks, \ours improves DINOv2 and DINOv3 by up to
31.7 CorLoc on object discovery, 30.2 maxF on saliency detection and
59.9~$\mathcal{J}$ on video object segmentation.

To further investigate what makes a model pareidolic, we trace its pareidolia index block
by block, which measures how much of the noise response recurs across noise images.
The trace reveals that all the ViTs we examine are strongly pareidolic in their
early and middle blocks.
In the last blocks, DINO becomes far less pareidolic, whereas the models that add a
patch-level loss to its recipe remain strongly pareidolic.
Controlled training confirms that the loss decides: trained from scratch with the
same data, architecture and schedule, a ViT remains strongly pareidolic to its last
block under iBOT's patch-level loss but becomes far less pareidolic under DINO's
image-level loss alone, and removing the patch-level loss from a trained iBOT has
the same effect.
This matches how pareidolia arises in human perception, where what is perceived
combines expectations with sensory evidence and pareidolia appears when the
expectations outweigh weak evidence~\cite{corlett2019hallucinations,lhotka2023no}.
A patch-level loss rewards such expectations: from the visible context, the model
must predict the features of masked patches whose content its input does not
contain, so it learns strong expectations about patch content, which then surface
even on noise.
DINO's image-level loss acts only on the \cls token, asking the outputs for
different crops of the same image to agree, so the patch features receive no target
of their own and are never trained to fill in content that is absent.

In summary, our contributions are as follows.
\begin{itemize}
  \item We identify and characterize pareidolia in modern self-supervised ViTs: on
  white noise, the leading directions of their features form a spatially coherent
  template, faint in DINO and strong in DINOv2 and DINOv3.
  \item We show that the pareidolic subspace of DINOv2 and DINOv3 holds most of the
  feature variance on natural images and hinders object emergence rather than
  destroying it.
  \item We propose \oursfullbold (\ours), a simple yet effective training-free
  calibration that scales down the pareidolic directions, measured on a single noise
  input, and recovers object emergence across diverse tasks.
  \item We trace pareidolia to the training objective: it forms in the middle blocks
  under both losses but stays strong to the last block only under a patch-level
  loss.
\end{itemize}

\section{Related Work}
\label{sec:related}

\paragraph{Pareidolia in humans and machines.}
Human pareidolia arises when learned expectations dominate weak or ambiguous
input~\cite{liu2014seeing,lhotka2023no}, in line with accounts of hallucination as
the work of overly strong priors~\cite{corlett2019hallucinations}; it engages
face-selective cortex~\cite{hadjikhani2009early,liu2014seeing,wardle2020rapid}, and
macaques share it~\cite{taubert2017face}.
It is studied with pure noise: observers led to expect faces or letters report them
on about a third of pure-noise images~\cite{liu2014seeing,rieth2011faces}, averaging the noise
that observers respond to recovers the internal representation behind their
reports~\cite{ahumada1971stimulus,gosselin2003superstitious}, an estimate of the template against which
they match the input~\cite{murray2011classification}, and a clinical noise
pareidolia test measures the tendency in patients~\cite{mamiya2016pareidolia}.
In machine vision, recent work tests whether networks share human face pareidolia
on photographs of illusory faces~\cite{hamilton2024seeing,gupta2025human,chen2026when},
and white-noise classification images reveal the class templates of
classifiers~\cite{borji2020white}.
The pareidolia we find is of another kind: a template of the model's own that the
dense features of self-supervised ViTs impose even on pure noise, which hinders
their object emergence and can be calibrated without training.

\vspace{-6pt}
\paragraph{Object emergence and self-supervised ViTs.}
DINO~\cite{caron2021emerging} showed that the \cls attention of a self-supervised
ViT segments objects without supervision, and a line of training-free methods reads
objects from the patch affinities of its final
layer~\cite{simeoni2021localizing,wang2022self,wang2023tokencut,wang2023cut,simeoni2023unsupervised,melaskyriazi2022deep,shin2022unsupervised,amir2022deep},
which defines the standard tests of object emergence; other methods train on these
affinities for semantic segmentation~\cite{hamilton2022unsupervised} or
object-centric learning~\cite{seitzer2023bridging}.
These methods were designed on DINO, while self-supervision has since moved to
patch-level objectives: masked image modeling predicts the pixels or features of
masked
patches~\cite{bao2022beit,xie2022simmim,he2022masked,baevski2022data2vec,assran2023self,darcet2025cluster},
iBOT~\cite{zhou2022ibot} adds masked patch prediction to DINO's self-distillation,
and DINOv2~\cite{oquab2024dinov2} and DINOv3~\cite{simeoni2025dinov3} scale it to
large curated data and lead dense benchmarks solved by trained heads.
Park et al.~\cite{park2023what} and Walmer et al.~\cite{walmer2023teaching} find
that masked modeling~\cite{xie2023revealing} keeps late attention local, whereas image-level objectives~\cite{caron2021emerging,chen2021empirical,assran2022masked}
make it homogeneous across queries; we instead compare the objectives by whether
the features stay pareidolic, and isolate the loss by training models that differ
only in it.

\section{White Noise Reveals Pareidolia}
\label{sec:probe}

In studies of human perception, pure noise reveals pareidolia because it offers no
evidence, so whatever pattern is reported comes from the observer.
We run the same test on frozen ViTs and ask whether their features impose a
template when the input has none.
We study three model families, DINO~\cite{caron2021emerging},
DINOv2~\cite{oquab2024dinov2} and DINOv3~\cite{simeoni2025dinov3}, with eight models
in total: DINO ViT-S/16 and B/16, DINOv2 ViT-S/14, B/14 and L/14, and DINOv3
ViT-S/16, B/16 and L/16.
For the analyses on natural images in \cref{sec:probe,sec:hinders,sec:discussion},
we use the first 500 images of VOC07 trainval.
We find pareidolia in other self-supervised ViTs as well, such as
iBOT~\cite{zhou2022ibot}, MAE~\cite{he2022masked} and
I-JEPA~\cite{assran2023self}.\footnote{Supplementary \cref{sec:supp-kappa} measures it
in 24 models from 11 families.}

\subsection{The noise test}
\label{sec:measure}

\paragraph{Probe.}
We feed each frozen model one white-noise image of $448\times448$ pixels, each
channel of each pixel drawn independently as
$\operatorname{clamp}(0.2\,z+0.5,\,0,\,1)$ with $z\sim\mathcal{N}(0,1)$, and collect
the final-layer keys of its $N_0$ patch tokens ($N_0=32^2$ at patch size 14 and
$28^2$ at 16) as the rows of $F_0\in\mathbb{R}^{N_0\times d}$.\footnote{Following
LOST~\cite{simeoni2021localizing} and TokenCut~\cite{wang2022self}, we use the
final-layer keys; results on the final outputs are given in supplementary
\cref{sec:supp-results}.}
With $\mu_0\in\mathbb{R}^d$ the mean of these rows, the SVD of the centred
response,
\begin{equation}
  F_0-\mathbf{1}\mu_0^\top=UDV^\top,\qquad \lambda_k=\sigma_k^2/N_0,
  \label{eq:svd}
\end{equation}
gives orthonormal directions $v_1,\dots,v_d$, the columns of $V$, ordered by the
singular values $\sigma_1\geq\sigma_2\geq\dots$ on the diagonal of $D$.
The coefficients of the $N_0$ patches along $v_k$ are $c^{(k)}=\sigma_k u_k$, with
$u_k$ the $k$-th column of $U$; their squares sum to $\sigma_k^2$, so $\lambda_k$ is
the variance of the response along $v_k$.

\vspace{-9pt}
\paragraph{Spatial coherence.}
Placed on the patch grid, $c^{(k)}$ forms a map, with $c^{(k)}_a$ the coefficient
of patch $a$.
The pixels of the input are independent and the patches do not overlap, so any
spatial correlation in these maps is contributed by the model rather than by the
input.
We measure it with Moran's $I$~\cite{moran1950notes}, which compares the
coefficient of each patch with those of its neighbours,
\begin{gather}
  I_k=\frac{N_0}{W}\,
  \frac{\sum_{a,b}w_{ab}\,c^{(k)}_a c^{(k)}_b}{\sum_a\big(c^{(k)}_a\big)^2},
  \label{eq:moran}\\
  z_k=\frac{I_k+1/(N_0-1)}{\sqrt{1/N_0}},
  \label{eq:moranz}
\end{gather}
where $w_{ab}=1$ if patches $a$ and $b$ share an edge of the grid and 0 otherwise,
and $W=\sum_{a,b}w_{ab}$.
$I_k$ approaches 1 when neighbouring patches take similar coefficients, and its
expectation for a map without spatial structure is $-1/(N_0-1)$; $z_k$ measures the
distance from this expectation in units of $1/\sqrt{N_0}$.
We call $v_k$ spatially coherent if $z_k\geq3$, and keep the leading run of
coherent directions, $v_1$ to $v_s$; this choice is examined in \cref{sec:ablation}
and supplementary \cref{sec:supp-details}.

\vspace{-6pt}
\paragraph{Definition.}
Following the human test, we call the features of a model \emph{pareidolic} when, on
white noise, a few of their directions are spatially coherent: where the input
offers no evidence, the coherent pattern these directions form is imposed by the
features, and by analogy with the human observer we call it their \emph{template}.
We call the span of these directions,
$\Vs=[v_1,\dots,v_s]\in\mathbb{R}^{d\times s}$, the \emph{pareidolic subspace}.
It appears without any image content, so its directions are a property of the model,
not of the input.

\subsection{How strongly the features are pareidolic}
\label{sec:share}

\paragraph{All three families impose a template.}
\Cref{fig:probe}a shows the response of one model per family, the Base size, to one
white-noise image.
In each, the leading directions form coherent spatial patterns, a weak
top-to-bottom gradient in DINO-B, grid-aligned patterns in DINOv2-B and smooth waves
across the image in DINOv3-B, whereas the first direction after the pareidolic
subspace shows no pattern.\footnote{More directions are shown in supplementary
\cref{fig:templates}.}
To compare how strongly the template weighs in each model, we use two measures, one
on noise and one on natural images.

\vspace{-6pt}
\paragraph{Pareidolia index.}
On noise, we measure how much of the response is the same template on every noise
image rather than a response to the particular noise.
We feed eight white-noise images with their own seeds, collect the keys of the
patch tokens, centre each response by its token mean to obtain
$R_1,\dots,R_8\in\mathbb{R}^{N_0\times d}$, and compute the \emph{pareidolia index}
(\cref{fig:probe}b)
\begin{equation}
  \kappa=\frac{\big\langle \tfrac14\sum_{i=1}^{4}R_i,\ \tfrac14\sum_{i=5}^{8}R_i\big\rangle_F}
              {\tfrac18\sum_{i=1}^{8}\lVert R_i\rVert_F^2},
  \label{eq:kappa}
\end{equation}
the inner product of two split-half means divided by the mean energy.
The two halves come from different noise images, so their inner product keeps only
what recurs regardless of the noise, and $\kappa$ is the share of the response
energy that recurs: $\kappa=1$ means that the keys respond identically to every
noise image, and $\kappa=0$ that the response changes entirely with the noise.
Unlike the rank rule, $\kappa$ needs no threshold, so it compares blocks and
models on one scale (\cref{sec:origin}).

\vspace{-6pt}
\paragraph{Excess share.}
On natural images, we measure how much of the variance of the features the
pareidolic subspace holds.
For one image, let $f_1,\dots,f_N$ be the features of its $N$ patches, centred by
their mean, $\Sigma=\frac1N\sum_i f_if_i^\top$ their covariance, and
$P_s=\Vs\Vs^\top$ the projection onto the subspace.
The subspace holds the share
$\rho_s=\operatorname{tr}(P_s\Sigma)/\operatorname{tr}\Sigma$ of the variance; since
the mean squared distance between two patch features is $2\operatorname{tr}\Sigma$,
$\rho_s$ is also the share of the squared distances between patches that lies in
the subspace.
Subspaces differ in size, and any $s$ of the $d$ directions would hold $s/d$ if the
variance were spread evenly, so we compare models by the \emph{excess share}
$\rho_s-s/d$, averaged over images (\cref{fig:probe}c).
The two measures agree: pareidolia is faint in DINO, while
strong in DINOv2 and DINOv3.

\section{Pareidolia Hinders Object Emergence}
\label{sec:hinders}

Object emergence, the hallmark ability of self-supervised ViTs to identify the
patches of an object on their own, relies on the similarities between patch
features.
In DINOv2 and DINOv3, the pareidolic subspace holds most of the feature variance and
therefore dominates these similarities.
We show that this is why both families score far below DINO: scaling the subspace
down closes most of the gap.

\subsection{Object emergence and its tests}
\label{sec:protocol}

We measure object emergence on its three standard tests: unsupervised single
object discovery, unsupervised saliency detection and unsupervised video object
segmentation.
We adopt TokenCut~\cite{wang2022self,wang2023tokencut}, which builds
a graph over the patches of an image from the cosine similarities of their
final-layer keys and partitions it with the Normalized
Cut~\cite{shi2000normalized}, and evaluate the eight models of \cref{sec:probe} at
native resolution.\footnote{More details are given in supplementary
\cref{sec:supp-details}.}

\begin{table*}[t]
  \centering
  \footnotesize
  \caption{\textbf{Unsupervised saliency detection} (\%, after the bilateral solver),
  with the gain of \ours over the raw features as a subscript.}
  \vspace{-6pt}
  \label{tab:saliency}
  \setlength{\tabcolsep}{10.5pt}
  \begin{tabular}{@{}lccccccccc@{\hspace{16pt}}}
    \toprule
    & \multicolumn{3}{c}{ECSSD} & \multicolumn{3}{c}{DUTS-TE} & \multicolumn{3}{c}{DUT-OMRON} \\
    \cmidrule(lr){2-4}\cmidrule(lr){5-7}\cmidrule(l){8-10}
    Model & maxF & IoU & Acc & maxF & IoU & Acc & maxF & IoU & Acc \\
    \midrule
    DINOv2-S/14 & 57.2 & 52.1 & 74.5 & 46.5 & 39.6 & 74.6 & 32.9 & 28.5 & 67.3 \\
    \quad + \ours & 79.4\gain{22.2} & 72.0\gain{19.9} & 89.3\gain{14.8} & 73.6\gain{27.1} & 62.2\gain{22.6} & 90.4\gain{15.8} & 63.1\gain{30.2} & 56.2\gain{27.7} & 85.7\gain{18.4} \\
    DINOv2-B/14 & 66.4 & 60.2 & 79.8 & 49.8 & 42.1 & 76.3 & 38.4 & 32.2 & 69.6 \\
    \quad + \ours & 81.2\gain{14.8} & 73.8\gain{13.6} & 89.7\gain{9.9} & 73.4\gain{23.6} & 62.2\gain{20.1} & 89.7\gain{13.4} & 62.5\gain{24.1} & 55.4\gain{23.2} & 85.2\gain{15.6} \\
    DINOv2-L/14 & 64.7 & 58.3 & 80.9 & 49.7 & 41.5 & 79.1 & 35.7 & 28.8 & 71.7 \\
    \quad + \ours & 81.3\gain{16.6} & 73.4\gain{15.1} & 90.1\gain{9.2} & 71.8\gain{22.1} & 60.7\gain{19.2} & 89.3\gain{10.2} & 60.8\gain{25.1} & 53.6\gain{24.8} & 85.2\gain{13.5} \\
    \midrule
    DINOv3-S/16 & 65.4 & 54.1 & 79.9 & 52.7 & 38.4 & 76.6 & 41.0 & 28.0 & 67.6 \\
    \quad + \ours & 77.8\gain{12.4} & 70.7\gain{16.6} & 89.1\gain{9.2} & 67.0\gain{14.3} & 55.9\gain{17.5} & 88.3\gain{11.7} & 57.9\gain{16.9} & 51.9\gain{23.9} & 84.5\gain{16.9} \\
    DINOv3-B/16 & 68.1 & 61.2 & 82.0 & 55.3 & 46.4 & 79.2 & 37.8 & 31.0 & 68.2 \\
    \quad + \ours & 81.9\gain{13.8} & 74.7\gain{13.5} & 90.3\gain{8.3} & 72.2\gain{16.9} & 60.6\gain{14.2} & 89.5\gain{10.3} & 61.8\gain{24.0} & 55.3\gain{24.3} & 85.5\gain{17.3} \\
    DINOv3-L/16 & 69.2 & 58.3 & 82.7 & 54.9 & 40.8 & 78.5 & 41.6 & 29.5 & 68.9 \\
    \quad + \ours & 73.0\gain{3.8} & 62.4\gain{4.1} & 87.9\gain{5.2} & 64.1\gain{9.2} & 51.1\gain{10.3} & 88.4\gain{9.9} & 48.9\gain{7.3} & 42.1\gain{12.6} & 83.8\gain{14.9} \\
    \bottomrule
  \end{tabular}
  \vspace{-9pt}
\end{table*}

\begin{table}[t]
  \centering
  \footnotesize
  \caption{\textbf{Unsupervised single object discovery} (CorLoc, \%), with the gain
  of \ours over the raw features as a subscript.}
  \vspace{-6pt}
  \label{tab:discovery}
  \setlength{\tabcolsep}{9pt}
  \begin{tabular}{@{}lccc@{\hspace{16pt}}}
    \toprule
    Model & VOC07 & VOC12 & COCO20K \\
    \midrule
    DINOv2-S/14 & 40.7 & 45.9 & 36.4 \\
    \quad + \ours & 63.4\gain{22.7} & 68.5\gain{22.6} & 54.3\gain{17.9} \\
    DINOv2-B/14 & 47.7 & 53.8 & 40.4 \\
    \quad + \ours & 65.4\gain{17.7} & 69.7\gain{15.9} & 55.6\gain{15.2} \\
    DINOv2-L/14 & 46.4 & 51.7 & 38.8 \\
    \quad + \ours & 65.5\gain{19.1} & 70.2\gain{18.5} & 57.2\gain{18.4} \\
    \midrule
    DINOv3-S/16 & 39.6 & 46.0 & 29.0 \\
    \quad + \ours & 65.4\gain{25.8} & 70.6\gain{24.6} & 54.2\gain{25.2} \\
    DINOv3-B/16 & 42.5 & 47.5 & 27.0 \\
    \quad + \ours & 68.1\gain{25.6} & 73.4\gain{25.9} & 58.7\gain{31.7} \\
    DINOv3-L/16 & 39.9 & 45.7 & 27.2 \\
    \quad + \ours & 63.2\gain{23.3} & 66.2\gain{20.5} & 54.5\gain{27.3} \\
    \bottomrule
  \end{tabular}
  \vspace{-9pt}
\end{table}

\vspace{-3pt}
\subsection{Object emergence fades}
\label{sec:anomaly}

As shown in \cref{tab:discovery,tab:saliency,tab:video} and \cref{fig:radar}, on
every task, DINOv2 and DINOv3 score far below DINO.\footnote{The performance of DINO
is given in supplementary \cref{sec:supp-dino}.}
These are also the two families whose pareidolia is strong (\cref{sec:share}).

\vspace{-3pt}
\subsection{Why does the excess hinder object emergence?}
\label{sec:why}

The training-free methods compare patches by the cosine similarity of their
features, which weighs every direction equally.
The squared distance between two patch features splits exactly into a part inside
the pareidolic subspace and a part outside it, and summed over all pairs of
patches, the part inside is the share $\rho_s$ of the whole.
When the subspace holds most of the variance, most of the difference between two
patches lies inside it, so the subspace, rather than the objects, decides which
patches look alike.

\begin{table}[t]
  \centering
  \footnotesize
  \caption{\textbf{Unsupervised video object segmentation} ($\mathcal{J}$, \%), with
  the gain of \ours over the raw features as a subscript.}
  \vspace{-6pt}
  \label{tab:video}
  \setlength{\tabcolsep}{9pt}
  \begin{tabular}{@{}lccc@{\hspace{16pt}}}
    \toprule
    Model & DAVIS-2016 & FBMS & SegTrackv2 \\
    \midrule
    DINOv2-S/14 & 23.7 & 35.5 & 31.1 \\
    \quad + \ours & 69.6\gain{45.9} & 75.5\gain{40.0} & 71.4\gain{40.3} \\
    DINOv2-B/14 & 43.7 & 59.0 & 62.2 \\
    \quad + \ours & 71.1\gain{27.4} & 76.7\gain{17.7} & 71.1\gain{8.9} \\
    DINOv2-L/14 & 51.8 & 59.3 & 56.4 \\
    \quad + \ours & 72.1\gain{20.3} & 71.9\gain{12.6} & 71.1\gain{14.7} \\
    \midrule
    DINOv3-S/16 & 15.6 & 19.2 & 10.6 \\
    \quad + \ours & 71.4\gain{55.8} & 72.4\gain{53.2} & 68.5\gain{57.9} \\
    DINOv3-B/16 & 22.9 & 27.5 & 20.2 \\
    \quad + \ours & 77.0\gain{54.1} & 73.5\gain{46.0} & 71.4\gain{51.2} \\
    DINOv3-L/16 & 18.6 & 24.1 & 15.8 \\
    \quad + \ours & 78.5\gain{59.9} & 65.8\gain{41.7} & 67.3\gain{51.5} \\
    \bottomrule
  \end{tabular}
  \vspace{-9pt}
\end{table}

\subsection{Scaling down the pareidolic subspace can restore object discovery}
\label{sec:rescale}

\paragraph{Experiment.}
In DINOv2-B and DINOv3-B, we scale the coordinates of the final keys along $\Vs$ by
a constant factor $\gamma\in[0,4]$, which moves the subspace's share of the
variance from zero ($\gamma=0$) through its raw value ($\gamma=1$) to above it, and
run object discovery at each $\gamma$ (\cref{fig:scaling}).

\vspace{-6pt}
\paragraph{Results.}
Object discovery follows the pareidolic subspace in both directions: scaling the
subspace down raises object discovery steadily in both models, with most of the
gain between $\gamma=1$ and $\gamma=0.5$, and scaling it up lowers it further.
Since scaling the subspace down brings the objects back, the object information is
still in the features: object emergence in DINOv2 and DINOv3 is hindered by
pareidolia rather than lost.

\begin{figure*}[t]
  \centering
  \includegraphics[width=0.96\textwidth]{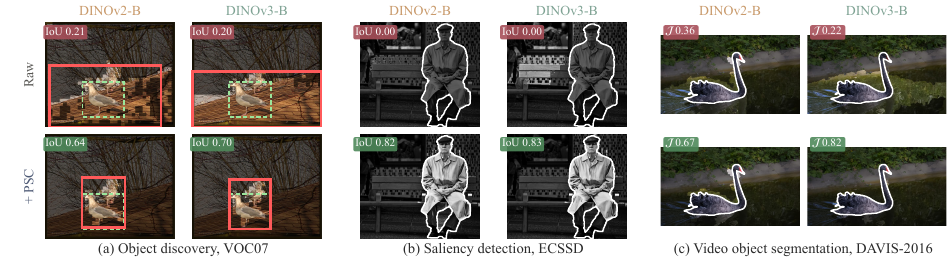}
  \vspace{-6pt}
  \caption{\textbf{Objects return once the pareidolic subspace is calibrated.} Rows:
  raw features and after \ours; in each task, DINOv2-B and DINOv3-B on the same image.
  In every tile, the region the method selects is bright and the rest dimmed, and the
  corner gives the IoU with the ground truth ($\mathcal{J}$ for video), green when at
  least 0.5. (a) Object discovery on VOC07: predicted box in red, ground truth of the
  object in question dashed in green, on an image drawn at random among those on which
  both raw models miss with a half-plane box (a box along two image borders) and both
  calibrated models localize the object. (b) Saliency detection on ECSSD, ground truth
  as a white contour, on an image drawn at random among those on which the IoU of both
  models rises above 0.5 after calibration. (c) Video object segmentation on
  DAVIS-2016: the middle annotated frame of the sequence at the median gain.}
  \vspace{-13pt}
  \label{fig:qualitative}
\end{figure*}

\section{\oursfull}
\label{sec:method}

The findings of \cref{sec:probe,sec:hinders} naturally lead to a training-free
calibration.
Because the pareidolic subspace is a property of the model, a single noise input
reveals its directions together with the excess variance they carry over the
remaining directions.
On natural images, this excess lets the subspace dominate the similarities between
patches, and scaling the subspace down brings the objects back.
We therefore propose \oursfullbold (\ours), which measures the pareidolic subspace only
on a noise input and then calibrates the features of natural inputs by scaling its
directions down to the variance level of the remaining directions.

\vspace{-6pt}
\paragraph{Measuring the subspace.}
The basis $\Vs$ and the noise spectrum $\lambda_1\geq\lambda_2\geq\dots$ come from
the single SVD of \cref{eq:svd} and the Moran rule of \cref{sec:measure}, so no
further forward pass is needed.
On the noise image, every direction of $\Vs$ carries far more variance than the
remaining directions.

\vspace{-6pt}
\paragraph{Calibrating the variance.}
\ours scales the coordinate of a key $f\in\mathbb{R}^d$ along each $v_j$, $j\leq s$,
by
\begin{equation}
  \alpha_j=\Big(\frac{\operatorname{median}_{k>s}\lambda_k}{\lambda_j}\Big)^{1/2},
  \label{eq:alpha}
\end{equation}
where the median runs over the nonzero $\lambda_k$ (the response of $N_0$ patches
has rank below $d$ when $N_0\le d$) and $\alpha_j$ is clipped to $[10^{-4},1]$.
On the noise image, each direction of $\Vs$ then carries the median variance of the
remaining directions, and the map leaves every other coordinate unchanged:
\begin{equation}
  \tilde f=f-\Vs\,\mathrm{diag}(\mathbf{1}-\alpha)\,\Vs^\top f .
  \label{eq:evc}
\end{equation}
The same $\Vs$ and $\alpha$, fixed once per model, calibrate the keys of every
image, and the downstream method then runs on $\tilde f$ exactly as it would
on $f$.

\vspace{-6pt}
\paragraph{Properties.}
Three properties follow from the construction: first, the map changes only the
variance of the pareidolic subspace, the quantity that the findings implicate, and
leaves the remaining $d-s$ coordinates untouched.
Second, it is invertible,
\begin{equation}
\vspace{-2pt}
  f=\tilde f+\Vs\,\mathrm{diag}\big(\alpha^{-1}-\mathbf{1}\big)\,\Vs^\top\tilde f ,
  \label{eq:inverse}
\end{equation}
so it discards nothing the features carry.
Third, it has no tuned parameter: $\Vs$ and $\alpha$ both come from the SVD
in \cref{eq:svd}.

\vspace{-6pt}
\paragraph{Cost and scope.}
\ours acts on the final keys that the training-free methods compare, and these
methods then run unchanged on $\tilde f$.
It costs one forward pass and one SVD per model, plus one rank-$s$ product per
token.
Since the pareidolic subspace is fixed by the model rather than by the input, one
basis serves inputs of every size.
We provide ablation studies on the noise input in \cref{sec:ablation}.

\section{Experiments}
\label{sec:exp}

\subsection{Setup}
\label{sec:setup}

We evaluate \ours on the six DINOv2 and DINOv3 models of \cref{sec:probe}, all
frozen, and follow the TokenCut protocol of each task.
For each model, the pareidolic subspace and its calibration scales are computed once
from a single noise image and applied unchanged to every image of every dataset;
for video, the noise image has the frame size.\footnote{More details are given in
supplementary \cref{sec:supp-details}.}

\subsection{Results}
\label{sec:results}

\ours improves every DINOv2 and DINOv3 model on all nine datasets and all metrics
(\cref{tab:discovery,tab:saliency,tab:video}), with the largest gains in video
object segmentation, where the raw DINOv3 models fail most.
The gains are as large on VOC12 and COCO20K, on which no threshold was selected, as
on VOC07.
\Cref{fig:qualitative} shows the three tasks on individual images: the raw keys of
DINOv2-B and DINOv3-B miss the object, and the calibrated keys return it.\footnote{More
details on the results of each task and a comparison with existing methods are given
in supplementary \cref{sec:supp-results,sec:supp-paradigms}.}

\section{Discussion and Ablations}
\label{sec:discussion}

Scaling the pareidolic subspace shows that pareidolia hinders object emergence in
DINOv2 and DINOv3, but two questions remain open: does the template alone produce
the failure in a model whose objects are intact, and what makes a model pareidolic?

\subsection{The template alone reproduces the failure}
\label{sec:inject}

\begin{figure}[t]
  \centering
  \vspace{-3pt}
  \includegraphics[width=0.9\linewidth]{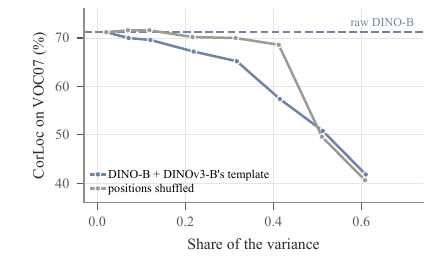}
  \vspace{-12pt}
  \caption{\textbf{DINOv3's template takes DINO's objects away.} Object discovery on VOC07 images when DINOv3-B's template is added to DINO-B's final keys at
  increasing shares of their variance (blue), and with its positions shuffled
  (grey). Dashed line is raw DINO-B.}
  \vspace{-15pt}
  \label{fig:inject}
\end{figure}

\paragraph{Experiment.}
We take DINOv3-B's template, its noise response within the pareidolic subspace,
resample it to the patch grid of each image, and add it to the final keys of DINO-B
at a prescribed share of their variance (\cref{fig:inject}).
This transfer is possible because the template is a property of the model rather
than of the image.
The probe finds the template through its spatial coherence, but on natural images
it competes through its share of the variance.
To tell the two apart, we also add it with the positions of its coefficients
shuffled across patches, which keeps its directions and its share but removes its
spatial pattern.

\vspace{-6pt}
\paragraph{Results.}
As the injected share grows, DINO-B's object discovery falls, and at a share close
to DINOv3-B, it drops to the level of DINOv3-B itself: the template alone
reproduces DINOv3's failure in a model whose objects are intact.
At this share, the injection with shuffled positions harms as much, thus what decides
the damage is how much of the variance the pareidolic subspace takes, not the
pattern it draws.

\subsection{Where pareidolia comes from}
\label{sec:origin}

\begin{figure}[t]
  \centering
  \vspace{-3pt}
  \includegraphics[width=0.9\linewidth]{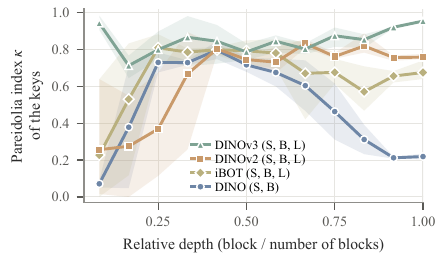}
  \vspace{-12pt}
  \caption{\textbf{All four families are strongly pareidolic in their middle blocks;
  only DINO becomes far less pareidolic by its last block.} Pareidolia index
  $\kappa$ of the keys per block
  (\cref{eq:kappa}), averaged over the sizes of each family (bands: min to max over
  sizes). DINO drops to 0.22 in its last block; iBOT, which adds only a patch-level
  loss to DINO's recipe, stays at 0.68, DINOv2 at 0.76 and DINOv3 at 0.95.}
  \vspace{-15pt}
  \label{fig:layerwise}
\end{figure}

\paragraph{Pareidolia forms in the middle blocks and stays strong only under a
patch-level loss.}
We compute the pareidolia index of the keys at every block of four families, DINO,
iBOT~\cite{zhou2022ibot}, DINOv2 and DINOv3 (\cref{fig:layerwise}).
iBOT shares DINO's data, architecture, crops and \cls loss and adds one ingredient,
masked patch prediction, the patch-level loss that DINOv2 and DINOv3 also use.
All four families are strongly pareidolic in early and middle blocks.
DINO becomes far less pareidolic over its last blocks, whereas iBOT, DINOv2
and DINOv3 remain strongly pareidolic to the end, and iBOT's object discovery falls
well below DINO's.

\vspace{-6pt}
\paragraph{Generality.}
We further extend the analysis to 24 self-supervised models from 11 families and
find that the split between the two losses remains consistent at the last block
(supplementary \cref{tab:kappa27}): every model trained with a patch-level loss, such as MAE~\cite{he2022masked}, I-JEPA~\cite{assran2023self} or
data2vec~\cite{baevski2022data2vec}, is far more pareidolic than every model
trained with an image-level loss alone, such as MoCo~v3~\cite{chen2021empirical} or
MSN~\cite{assran2022masked}.
Released models, however, differ in more than their loss, so we isolate the loss by
controlled training.

\begin{figure*}[t]
  \centering
  \includegraphics[width=0.9\textwidth]{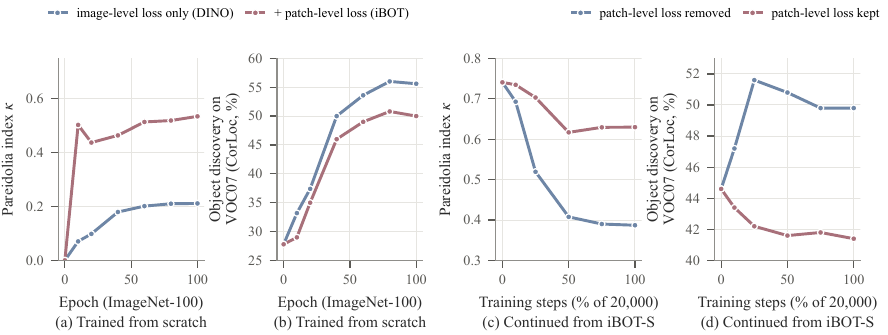}
  \vspace{-6pt}
  \caption{\textbf{The patch-level loss keeps pareidolia strong, and it costs object
  discovery.} ViT-S/16 pairs that differ only in the patch-level loss.
  (a, b) Trained from scratch on ImageNet-100 with DINO's image-level loss alone
  (blue) or with iBOT's patch-level loss added (rose): pareidolia index, and object
  discovery on VOC07 images, per epoch. (c, d) The released iBOT-S continued
  for 20,000 steps with its patch-level loss kept (rose) or removed (blue).}
  \vspace{-13pt}
  \label{fig:controlled}
\end{figure*}

\vspace{-6pt}
\paragraph{Changing only the loss.}
To isolate the objective, we trained pairs of ViT-S/16 that differ in nothing
else.
From scratch, on ImageNet-100~\cite{deng2009imagenet,tian2020contrastive} for 100 epochs with the same data order,
augmentation, schedule and seed, one model uses DINO's image-level loss alone and
the other adds iBOT's patch-level loss.
In the reverse direction, we take the released iBOT-S and continue it for 20,000
steps twice, once with its patch-level loss and once with the \cls loss
alone.\footnote{More details are given in supplementary \cref{sec:supp-controlled}.}

\vspace{-6pt}
\paragraph{Results.}
From scratch (\cref{fig:controlled}a), the model with the patch-level loss becomes
strongly pareidolic in its last block within 10 epochs, while the model with the
image-level loss alone stays far less pareidolic, a gap that never closes.
From the released iBOT-S (\cref{fig:controlled}c), removing the patch-level loss
makes the model far less pareidolic in its last block within the first half of
training, while the model that keeps the loss stays strongly pareidolic.
The objective also affects object discovery (\cref{fig:controlled}b, d): in both
pairs, the model with the patch-level loss is worse at object discovery, and in the
second pair the gap opens as the pareidolia weakens, while its $k$-NN accuracy on
ImageNet-100 is equal or higher.
The patch-level loss therefore does not weaken the representation; it keeps the
representation pareidolic.

\vspace{-6pt}
\paragraph{Reading.}
The patch-level loss keeps pareidolia strong to the last block; image-level
training does not prevent it from forming but lets it weaken.
This matches how pareidolia arises in human perception, where what is perceived
combines expectations with sensory evidence and pareidolia appears when the
expectations outweigh weak or ambiguous evidence~\cite{liu2014seeing,lhotka2023no};
overly strong expectations can even produce a percept with no input at
all~\cite{corlett2019hallucinations}.
A patch-level loss rewards strong expectations: each masked patch must be predicted
from its visible context, with no evidence of its own, so the patch features learn
what a patch is expected to contain, and white noise, which offers no evidence
anywhere, exposes these expectations as a template.
DINO's image-level loss acts only on the \cls token, asking the outputs for
different crops of the same image to agree, so the patch features receive no target
of their own and are never rewarded for such expectations.\footnote{More details on
the template are given in supplementary \cref{sec:supp-discussion}.}

\subsection{Ablations}
\label{sec:ablation}

\begin{table}[t]
  \centering
  \footnotesize
  \caption{\textbf{Ablations} (object discovery on full VOC07, CorLoc \%). Defaults
  in grey; $s$ is the rank the Moran rule selects on each noise image; $\pm$ is the
  std over five seeds or draws. In (c), raw DINOv2-B scores
  47.7 CorLoc with 14-pixel patches and 44.4 with 16.}
  \vspace{-6pt}
  \label{tab:ablation}
  \setlength{\tabcolsep}{4pt}
  \begin{tabular}{@{}lcc@{}}
    \toprule
    Variant & DINOv2-B & DINOv3-B \\
    \midrule
    \multicolumn{3}{@{}l}{\emph{(a) Noise image}} \\
    \gray{white noise, $448^2$ ($s=62$ / 13)} & \gray{65.4} & \gray{68.1} \\
    white noise, 5 other seeds & 64.9 $\pm$ 0.7 & 68.0 $\pm$ 0.2 \\
    uniform noise ($s=62$ / 13) & 66.4 & 67.8 \\
    pink noise ($s=119$ / 125) & 62.6 & 64.5 \\
    constant gray ($s=254$ / 256) & 66.1 & 60.4 \\
    \midrule
    \multicolumn{3}{@{}l}{\emph{(b) Calibrated directions}} \\
    \gray{Moran rule ($s=62$ / 13)} & \gray{65.4} & \gray{68.1} \\
    first 4 & 43.2 & 65.2 \\
    first 16 & 59.4 & 67.7 \\
    first 64 & 65.0 & 62.2 \\
    $s$ random directions (5 draws) & 39.9 $\pm$ 0.6 & 42.6 $\pm$ 0.3 \\
    \midrule
    \multicolumn{3}{@{}l}{\emph{(c) Patch size}} \\
    \gray{14, released ($s=62$)} & \gray{65.4} & -- \\
    16, patch projection resized ($s=61$) & 63.8 & -- \\
    \bottomrule
  \end{tabular}
  \vspace{-3pt}
\end{table}

We ablate the \textbf{noise image} and the \textbf{calibrated directions} on
DINOv2-B and DINOv3-B, and the \textbf{patch size} on DINOv2-B
(\cref{tab:ablation}).
Other white-noise seeds and uniform noise barely change object discovery, whereas
on a constant gray image the Moran rule accepts hundreds of directions and object
discovery drops on DINOv3-B.
A fixed number of leading directions either misses part of the subspace or reaches
directions that carry content, and $s$ random directions change nothing, so the
gain comes from which directions are calibrated.
Resizing DINOv2-B's patch projection from 14 to 16 pixels, the patch size of DINO
and DINOv3, leaves the pareidolic subspace and the recovery nearly unchanged, so
the pareidolia of DINOv2 does not come from its patch size.\footnote{More details
are given in supplementary \cref{sec:supp-ablation}.}

\section{Conclusion}
\label{sec:conclusion}

The features of modern self-supervised ViTs are pareidolic: on white noise, which
contains neither objects nor spatial structure, a few of their directions form a
spatially coherent template.
In DINOv2 and DINOv3, the pareidolic subspace these directions span holds most of
the feature variance on natural images, and their object emergence is hindered by
it rather than lost.
\ours measures this subspace only on a noise input, scales its directions down
without training, and recovers object emergence across diverse tasks.
Pareidolia forms in the middle blocks of every self-supervised model we examine but
stays strong to the last block only in models trained with a patch-level loss.

\section*{LLM Usage Statement}
We certify the use of AI to aid in the writing, polishing, and figure plotting of this manuscript, and the use of AI to aid code implementations of involved experiments in this work.
All AI-assisted content shown in this manuscript has been reviewed and verified by the authors.
We take responsibility for the final content of this work, including text, claims or artifacts produced with the aid of AI.

{\small
\bibliographystyle{meta/ieeenat_fullname}
\bibliography{sections/11_references}
}

\clearpage\appendix
\raggedbottom
\renewcommand{\thesection}{\Alph{section}}
\renewcommand{\thetable}{\Alph{table}}
\renewcommand{\thefigure}{\Alph{figure}}
\renewcommand{\theequation}{\Alph{equation}}
\renewcommand{\theHtable}{supp.\arabic{table}}
\renewcommand{\theHfigure}{supp.\arabic{figure}}
\renewcommand{\theHequation}{supp.\arabic{equation}}
\setcounter{table}{0}
\setcounter{figure}{0}
\setcounter{equation}{0}
\maketitlesupplementary
\addtocontents{toc}{\protect\setcounter{tocdepth}{2}}
\def\supptocgobble#1\hskip-1em.{#1}
\renewcommand{\cftsecaftersnumb}{\supptocgobble}
\renewcommand{\cftsubsecaftersnumb}{\supptocgobble}
\renewcommand{\cftsecfont}{\normalfont}
\renewcommand{\cftsecpagefont}{\normalfont}
\renewcommand{\cftsubsecfont}{\normalfont}
\renewcommand{\cftsubsecpagefont}{\normalfont}
\setlength{\cftsubsecindent}{\cftsecnumwidth}
\renewcommand{\cftdotsep}{1}
\renewcommand{\cftsecleader}{\cftdotfill{\cftdotsep}}
\renewcommand{\cftsubsecleader}{\cftdotfill{\cftdotsep}}
\cftsetpnumwidth{1.5em}
\cftsetrmarg{1.5em plus 1fil}
\renewcommand{\cfttoctitlefont}{\Large\bfseries}
\setlength{\cftbeforetoctitleskip}{0pt}
\setlength{\cftaftertoctitleskip}{8pt}
\setlength{\cftbeforesecskip}{1pt}
\tableofcontents
\section{Additional Experimental Details}
\label{sec:supp-details}

\paragraph{Models.}
We evaluate DINO~\cite{caron2021emerging} ViT-S/16 and ViT-B/16,
DINOv2~\cite{oquab2024dinov2} ViT-S/14, ViT-B/14 and ViT-L/14, and
DINOv3~\cite{simeoni2025dinov3} ViT-S/16, ViT-B/16 and ViT-L/16.
For each model, the pareidolic subspace and its calibration scales are computed
once from a single noise image (\cref{sec:method}) and applied unchanged to every
image of every dataset; for video, whose frames are resized to a fixed size, the
noise image has the frame size.

\paragraph{Protocol.}
We follow the setting of TokenCut~\cite{wang2022self,wang2023tokencut} for each
task.
TokenCut builds a graph over the patches of an image from the cosine similarities
of their final-layer keys, assigning weight 1 to edges above a threshold $\tau$ and
$10^{-5}$ to the others, and splits it with the Normalized
Cut~\cite{shi2000normalized}.
For object discovery, images are processed at native resolution, and we report
CorLoc, the fraction of images whose predicted box overlaps a ground-truth box with
an IoU of at least 0.5, on the trainval sets of VOC07 and
VOC12~\cite{everingham2010pascal} and on COCO20K~\cite{lin2014microsoft,vo2020toward}.
For saliency detection, we report the maximal F-measure (maxF), IoU and accuracy
after the bilateral solver~\cite{barron2016fast} on ECSSD~\cite{shi2016hierarchical},
DUTS-TE~\cite{wang2017learning} and DUT-OMRON~\cite{yang2013saliency}.
For video object segmentation, frames are resized to $320\times576$, the graph
averages the key similarities of the frames and of their RAFT optical
flow~\cite{teed2020raft} rendered as images, masks are refined with a
CRF~\cite{krahenbuhl2011efficient}, and we report the Jaccard index $\mathcal{J}$ on
DAVIS-2016~\cite{perazzi2016benchmark} (20 validation videos),
FBMS~\cite{ochs2014segmentation} (30 test videos) and SegTrackv2~\cite{li2013video}
(14 videos).

\paragraph{Rank rule.}
The rank $s$ is the length of the leading run of noise directions with Moran's
$z_k\geq3$ (\cref{eq:moranz}), a choice that the ablation shows to be robust
(\cref{tab:ablation}b and \cref{sec:supp-ablation}).

\section{Additional Results on More Tasks}
\label{sec:supp-further}

\begin{table*}[t]
  \centering
  \footnotesize
  \caption{\textbf{\ours on further training-free tasks}, with the gain of \ours over
  the raw features as a subscript. \cls attention and CUB: CorLoc; k-means, LaVG and
  CASS: mIoU.}
  \vspace{-6pt}
  \label{tab:further}
  \setlength{\tabcolsep}{9pt}
  \begin{tabular}{@{}lcccccccc@{\hspace{14pt}}}
    \toprule
    & \multicolumn{3}{c}{\cls attention} & CUB & \multicolumn{2}{c}{k-means} & LaVG & CASS \\
    \cmidrule(lr){2-4}\cmidrule(lr){5-5}\cmidrule(lr){6-7}\cmidrule(lr){8-8}\cmidrule(l){9-9}
    Model & VOC07 & VOC12 & COCO20K & test & VOC12 & COCO-Stuff & VOC21 & VOC21 \\
    \midrule
    DINOv2-B/14 & 14.4 & 11.1 & 18.4 & 76.6 & 2.6 & 7.6 & 56.8 & 58.8 \\
    \quad + \ours & 48.8\gain{34.4} & 53.5\gain{42.4} & 35.6\gain{17.2} & 95.0\gain{18.4} & 9.2\gain{6.6} & 15.1\gain{7.5} & 62.4\gain{5.6} & 63.2\gain{4.4} \\
    \midrule
    DINOv3-B/16 & 24.0 & 23.0 & 24.2 & 78.1 & 2.3 & 3.3 & 44.8 & 61.9 \\
    \quad + \ours & 49.6\gain{25.6} & 55.1\gain{32.1} & 37.7\gain{13.5} & 95.0\gain{16.9} & 8.4\gain{6.1} & 13.1\gain{9.8} & 61.8\gain{17.0} & 65.8\gain{3.9} \\
    \bottomrule
  \end{tabular}
  \vspace{-6pt}
\end{table*}

\paragraph{Training-free evaluation.}
Beyond the three tasks of the main text, we evaluate DINOv2-B and DINOv3-B on further
training-free tasks, each with its released protocol (\cref{tab:further}).
The first reads objects from the \cls attention of the last block, in which object
emergence was first observed~\cite{caron2021emerging}: we recompute the attention of
the \cls query to the calibrated patch keys, average it over the heads, keep the
smallest set of patches that holds 60\% of the attention, and box its largest
connected component (CorLoc on VOC07, VOC12 and COCO20K).
The second is object discovery with TokenCut on the 5,794 test images of
CUB-200-2011~\cite{wah2011caltech}, a fine-grained dataset of birds (CorLoc).
The third segments VOC12 val and COCO-Stuff~\cite{caesar2018coco} by clustering the
final keys of all images with k-means into as many clusters as classes, 21 and 27,
matched to the classes by the Hungarian algorithm (mIoU), the training-free baseline
of STEGO~\cite{hamilton2022unsupervised}.
The last two segment VOC12 val by open-vocabulary classes, its 21 classes given as
text (mIoU): LaVG~\cite{kang2024defense} finds object masks with the Normalized Cut
of the features and names each mask with CLIP~\cite{radford2021learning}, and
CASS~\cite{kim2025distilling} distills the spectral structure of the features into the
attention of CLIP, so that its predictions follow objects.
\ours improves both models on every task and dataset, further confirming its
effectiveness and its generality across training-free tasks.

\begin{table}[t]
  \centering
  \footnotesize
  \caption{\textbf{\ours leaves supervised evaluation unchanged.} ADE20K: mIoU of a
  linear segmentation head with the protocol of DINOv2; ImageNet-1k: top-1 of linear
  and 20-NN classification with 100 training images per class.}
  \vspace{-6pt}
  \label{tab:heads}
  \setlength{\tabcolsep}{7pt}
  \begin{tabular}{@{}lccc@{}}
    \toprule
    & ADE20K & \multicolumn{2}{c}{ImageNet-1k} \\
    \cmidrule(lr){2-2}\cmidrule(l){3-4}
    Model & mIoU & linear & 20-NN \\
    \midrule
    DINOv3-B/16 & 51.7 & 77.2 & 69.9 \\
    \quad + \ours & 51.7 & 77.2 & 69.9 \\
    \bottomrule
  \end{tabular}
\end{table}

\paragraph{Supervised evaluation.}
\ours scales the pareidolic subspace down to the variance level of the remaining
directions, which removes the excess variance that lets the subspace dominate the
similarities between patches, and it does so at no cost to supervised training.
With and without \ours, DINOv3-B reaches the same mIoU on
ADE20K~\cite{zhou2017scene} with a linear segmentation head trained with the
protocol of DINOv2~\cite{oquab2024dinov2}, and the same top-1 accuracy on
ImageNet-1k~\cite{deng2009imagenet} in linear and 20-NN classification with 100
training images per class (\cref{tab:heads}).

\section{Additional Results on TokenCut Feature Selection}
\label{sec:supp-results}

\begin{table}[t]
  \centering
  \footnotesize
  \caption{\textbf{Object discovery on the final outputs} (CorLoc, \%), the tokens
  after the last block, with the gain of \ours over the raw features as a
  subscript.}
  \vspace{-6pt}
  \label{tab:outputs}
  \setlength{\tabcolsep}{9pt}
  \begin{tabular}{@{}lccc@{\hspace{16pt}}}
    \toprule
    Model & VOC07 & VOC12 & COCO20K \\
    \midrule
    DINOv2-S/14 & 51.7 & 56.4 & 45.0 \\
    \quad + \ours & 53.5\gain{1.8} & 57.7\gain{1.3} & 46.7\gain{1.7} \\
    DINOv2-B/14 & 42.1 & 48.3 & 35.1 \\
    \quad + \ours & 52.1\gain{10.0} & 56.4\gain{8.1} & 46.2\gain{11.1} \\
    DINOv2-L/14 & 36.8 & 41.7 & 28.3 \\
    \quad + \ours & 55.5\gain{18.7} & 60.0\gain{18.3} & 46.4\gain{18.1} \\
    \midrule
    DINOv3-S/16 & 40.7 & 46.5 & 33.9 \\
    \quad + \ours & 53.1\gain{12.4} & 58.2\gain{11.7} & 44.5\gain{10.6} \\
    DINOv3-B/16 & 47.5 & 52.4 & 36.9 \\
    \quad + \ours & 57.8\gain{10.3} & 62.1\gain{9.7} & 45.9\gain{9.0} \\
    DINOv3-L/16 & 48.6 & 52.9 & 38.0 \\
    \quad + \ours & 54.8\gain{6.2} & 60.4\gain{7.5} & 43.4\gain{5.4} \\
    \bottomrule
  \end{tabular}
  \vspace{-6pt}
\end{table}

Following LOST~\cite{simeoni2021localizing} and TokenCut~\cite{wang2022self}, the
main text runs TokenCut on the final-layer keys.
We additionally provide object discovery on the final outputs, the tokens after the
last block (\cref{tab:outputs}): \ours improves every DINOv2 and DINOv3 model there
as well, on all three datasets.
\ours is thus robust to the choice between the final keys and the final
outputs.

\vspace{-6pt}
\section{Additional Results on Comparison with Existing Paradigms}
\label{sec:supp-paradigms}

\paragraph{Related methods.}
Registers~\cite{darcet2024vision} remove the high-norm artifact tokens of DINOv2,
which resemble the attention sinks and massive activations of language
models~\cite{xiao2024efficient,sun2024massive}, and test-time
registers~\cite{jiang2025vision} move them into an extra token without retraining.
SINDER~\cite{wang2024sinder} traces such artifacts to singular defects of the network
and repairs them by fine-tuning.
DVT~\cite{yang2024denoising} learns a denoiser for position-dependent artifacts in
the output features and notes that ``the fundamental underlying reason behind this
property remains unclear''; RASA~\cite{venkataramanan2025franca} removes the
directions along which a linear regressor predicts each patch's 2D position; and
Pawlowsky et al.~\cite{pawlowsky2026what} show that ALiBi positional encoding
reduces positional bias in ViTs.
Each of these works targets a component named in advance, artifact tokens or
positional information, and Object-DINO~\cite{rawlekar2026finding} aggregates the
object information spread over the queries, keys and values of many heads and
layers.
We instead probe the features with an input that has no content, so that whatever
structure they impose shows without being named.
Registers act on a few tokens, while the pareidolic subspace spans a few dense
directions; and whether a linear regressor can predict position is unchanged by
any invertible map of the features~\cite{hewitt2019designing,elazar2021amnesic},
whereas the similarities that training-free methods compare are not.
We add two controls: natural-image PCA removes as many leading principal directions
of natural-image keys as \ours calibrates, and global ZCA whitening whitens all
directions of the keys.

\begin{table*}[t]
  \centering
  \footnotesize
  \caption{\textbf{Comparison with existing paradigms} (object discovery, CorLoc \%).
  DINOv3 is trained with registers, and DVT is released for DINOv2 only.
  $^\dagger$For a fair comparison, we report the results of DVT on the final keys.}
  \vspace{-6pt}
  \label{tab:paradigms}
  \setlength{\tabcolsep}{3.3pt}
  \begin{tabular}{@{}lcccccc|cccccc|cccccc@{}}
    \toprule
    & \multicolumn{6}{c|}{VOC07} & \multicolumn{6}{c|}{VOC12} & \multicolumn{6}{c}{COCO20K} \\
    & \multicolumn{3}{c}{DINOv2} & \multicolumn{3}{c|}{DINOv3} & \multicolumn{3}{c}{DINOv2} & \multicolumn{3}{c|}{DINOv3} & \multicolumn{3}{c}{DINOv2} & \multicolumn{3}{c}{DINOv3} \\
    Method & S & B & L & S & B & L & S & B & L & S & B & L & S & B & L & S & B & L \\
    \midrule
    Raw final keys & 40.7 & 47.7 & 46.4 & 39.6 & 42.5 & 39.9 & 45.9 & 53.8 & 51.7 & 46.0 & 47.5 & 45.7 & 36.4 & 40.4 & 38.8 & 29.0 & 27.0 & 27.2 \\
    Natural-image PCA & 32.5 & 30.1 & 27.8 & 39.0 & 59.1 & 41.4 & 34.0 & 32.5 & 32.8 & 43.3 & 63.4 & 45.3 & 29.6 & 29.8 & 23.5 & 40.4 & 53.1 & 34.4 \\
    Global ZCA whitening & 44.9 & 36.4 & 39.4 & 37.5 & 40.8 & 42.9 & 49.9 & 41.9 & 42.6 & 43.4 & 46.4 & 47.8 & 40.0 & 27.8 & 35.9 & 31.5 & 35.7 & 34.6 \\
    Registers~\cite{darcet2024vision} & 43.3 & 49.0 & 58.9 & \multicolumn{3}{c|}{built in} & 47.9 & 54.7 & 63.0 & \multicolumn{3}{c|}{built in} & 36.7 & 41.9 & 48.2 & \multicolumn{3}{c}{built in} \\
    DVT$^\dagger$~\cite{yang2024denoising} & 42.5 & 40.2 & 44.1 & -- & -- & -- & 49.7 & 46.8 & 49.4 & -- & -- & -- & 37.4 & 32.1 & 37.3 & -- & -- & -- \\
    Test-time registers~\cite{jiang2025vision} & -- & -- & 49.2 & -- & -- & -- & -- & -- & 53.7 & -- & -- & -- & -- & -- & 41.9 & -- & -- & -- \\
    RASA~\cite{venkataramanan2025franca} & 45.3 & 49.6 & 49.2 & 45.3 & 56.0 & 52.6 & 51.3 & 55.4 & 54.4 & 52.2 & 61.2 & 57.3 & 39.9 & 42.2 & 41.1 & 32.5 & 39.6 & 34.9 \\
    Object-DINO~\cite{rawlekar2026finding} & 42.0 & 46.9 & 43.0 & 34.6 & 48.5 & 40.0 & 46.1 & 52.4 & 46.3 & 39.2 & 51.4 & 44.1 & 37.9 & 39.0 & 38.0 & 26.9 & 37.2 & 29.3 \\
    \midrule
    \ours & \textbf{63.4} & \textbf{65.4} & \textbf{65.5} & \textbf{65.4} & \textbf{68.1} & \textbf{63.2} & \textbf{68.5} & \textbf{69.7} & \textbf{70.2} & \textbf{70.6} & \textbf{73.4} & \textbf{66.2} & \textbf{54.3} & \textbf{55.6} & \textbf{57.2} & \textbf{54.2} & \textbf{58.7} & \textbf{54.5} \\
    \bottomrule
  \end{tabular}
  \vspace{-6pt}
\end{table*}

\paragraph{Setup.}
Every method is applied to the same models and read with TokenCut on the final
keys, except Object-DINO, which builds its affinity from the queries, keys and values
of all layers (\cref{tab:paradigms}).
Registers are evaluated with the DINOv2 checkpoints trained with them; DINOv3 is
trained with registers, so the comparison does not apply to it.
Test-time registers are available for DINOv2-L only, and DVT for DINOv2 only.

\paragraph{Results.}
\ours outperforms every existing method and both controls on every model and
dataset (\cref{tab:paradigms}).
On VOC07, it leads the best of them by 6.6 to 20.1 CorLoc.
Registers and RASA recover 1.3 to 13.5 points over the raw keys and Object-DINO
changes them by $-5.0$ to $+6.0$, whereas both controls fall below the raw keys on
four of the six models, so the gain of \ours comes from the directions selected
on noise, not from removing leading directions or whitening in general.

\section{Additional Results on DINO}
\label{sec:supp-dino}

\begin{table}[t]
  \centering
  \footnotesize
  \caption{\textbf{Performance of DINO} on the three tasks of the main text. Object
  discovery: CorLoc; saliency detection: maxF after the bilateral solver; video object
  segmentation: $\mathcal{J}$.}
  \vspace{-6pt}
  \label{tab:dino}
  \setlength{\tabcolsep}{6pt}
  \begin{tabular}{@{}llcc@{}}
    \toprule
    Task & Dataset & DINO-S/16 & DINO-B/16 \\
    \midrule
    Object discovery & VOC07 & 68.5 & 68.7 \\
    & VOC12 & 72.1 & 72.3 \\
    & COCO20K & 58.2 & 59.0 \\
    \midrule
    Saliency detection & ECSSD & 88.0 & 86.8 \\
    & DUTS-TE & 75.3 & 74.1 \\
    & DUT-OMRON & 68.9 & 67.5 \\
    \midrule
    Video segmentation & DAVIS-2016 & 78.4 & 77.6 \\
    & FBMS & 66.1 & 65.2 \\
    & SegTrackv2 & 69.3 & 69.1 \\
    \bottomrule
  \end{tabular}
  \vspace{-6pt}
\end{table}

\Cref{tab:dino} gives the results of DINO-S and DINO-B on the three tasks,
with the protocol of the main text.
We regard the performance of DINO as the target for DINOv2 and DINOv3, since DINO is
the model in which object emergence was first observed~\cite{caron2021emerging} and
its pareidolia is faint (\cref{sec:share}).

\section{Further Discussion on Ablation Studies}
\label{sec:supp-ablation}

\paragraph{Noise image.}
Every noise image we tried yields a large gain over the raw features, which reach
47.7 and 42.5 CorLoc on DINOv2-B and DINOv3-B (\cref{tab:ablation}a).
Other white-noise seeds and uniform noise change the result by at most 1.3 points,
and pink noise and a constant gray image, on which the Moran rule accepts more than
a hundred directions, still reach at least 60.4 CorLoc.
We use white noise, which gives the best result on DINOv3-B and is within 1.0
point of the best on DINOv2-B.

\paragraph{Number of calibrated directions.}
The rank selected by the Moran rule, 62 for DINOv2-B and 13 for DINOv3-B, gives
the best object discovery on both models (\cref{tab:ablation}b).
A fixed number of leading directions close to this rank comes within 0.4 points of
it, and numbers farther away give less, such as 43.2 CorLoc with the first 4
directions on DINOv2-B and 62.2 with the first 64 on DINOv3-B.
Calibrating $s$ random directions with the same scales does not recover object
discovery (39.9 and 42.6), so the gain comes from which directions are calibrated.

\paragraph{Patch size.}
To test whether the 14-pixel patches of DINOv2 matter, we resize the patch
projection of DINOv2-B bilinearly to $16\times16$ pixels and keep every other
weight, without retraining.
The pareidolic subspace stays nearly the same ($s=61$ against 62, pareidolia index
0.69 against 0.75, excess share 0.44 in both), and \ours raises object discovery
on VOC07 by 19.4 CorLoc, from 44.4 to 63.8, against 17.7 with 14-pixel patches, so
the patch size affects neither the pareidolic subspace nor the effectiveness of
\ours.

\section{Further Discussion on More Self-Supervised ViTs}
\label{sec:supp-kappa}

\begin{table}[t]
  \centering
  \footnotesize
  \caption{\textbf{Patch-level losses keep pareidolia strong to the last block.}
  Pareidolia index $\kappa$ of the final keys (\cref{eq:kappa}) for 24
  self-supervised models from 11 families, grouped by training objective.}
  \vspace{-6pt}
  \label{tab:kappa27}
  \setlength{\tabcolsep}{6pt}
  \begin{tabular}{@{}lc@{}}
    \toprule
    Model & $\kappa$ \\
    \midrule
    \multicolumn{2}{@{}l}{\emph{Image-level loss}} \\
    DINO-S/16, B/16~\cite{caron2021emerging} & 0.24, 0.20 \\
    MoCo~v3-B/16~\cite{chen2021empirical} & 0.08 \\
    MSN-B/16~\cite{assran2022masked} & 0.20 \\
    \midrule
    \multicolumn{2}{@{}l}{\emph{Patch-level loss}} \\
    iBOT-S/16, B/16, L/16~\cite{zhou2022ibot} & 0.74, 0.65, 0.64 \\
    MAE-B/16~\cite{he2022masked} & 0.67 \\
    DINOv2-S/14, B/14, L/14, g/14~\cite{oquab2024dinov2} & 0.75, 0.75, 0.78, 0.65 \\
    DINOv2 with registers, S/B/L/g~\cite{darcet2024vision} & 0.54, 0.58, 0.56, 0.48 \\
    DINOv3-S/16, B/16, L/16~\cite{simeoni2025dinov3} & 0.95, 0.96, 0.96 \\
    Franca-B/14, L/14~\cite{venkataramanan2025franca} & 0.79, 0.63 \\
    I-JEPA-H/14~\cite{assran2023self} & 0.54 \\
    CAPI-L/14~\cite{darcet2025cluster} & 0.54 \\
    data2vec-B/16~\cite{baevski2022data2vec} & 0.70 \\
    \bottomrule
  \end{tabular}
  \vspace{-6pt}
\end{table}

\begin{figure}[t]
  \centering
  \includegraphics[width=0.9\linewidth]{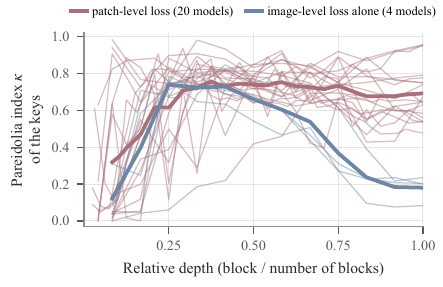}
  \vspace{-12pt}
  \caption{\textbf{Pareidolia rises alike under both losses through the early
  blocks; only the models trained with an image-level loss alone become far less
  pareidolic in the later blocks.} Pareidolia index $\kappa$ of the keys from the
  first block to the last, against relative depth, for the 24 models of
  \cref{tab:kappa27} (thin lines), and the mean of each loss group (bold).}
  \label{fig:kappa-blocks}
  \vspace{-12pt}
\end{figure}

Beyond the DINO lineage, we measure the pareidolia index of the final keys for 24
self-supervised ViTs from 11 families (\cref{tab:kappa27}) and find the same
pattern as in DINO, DINOv2 and DINOv3: models trained with a patch-level loss are
strongly pareidolic, with $\kappa$ from 0.48 to 0.96, whereas models trained with an
image-level loss alone are far less pareidolic, with $\kappa$ from 0.08 to 0.24.
The two ranges do not overlap.
\Cref{fig:kappa-blocks} follows $\kappa$ through the blocks.
Up to the middle blocks, the two groups follow the same trend: the mean $\kappa$ of
each group rises over the early blocks, the two means reach the same level, about
0.72, by a relative depth of 0.33, and they stay within 0.03 of each other up to 0.42.
The trends diverge only in the later blocks, where the mean of the models trained
with an image-level loss alone falls steadily, to 0.18 at the last block, whereas the
mean of the models trained with a patch-level loss stays between 0.68 and 0.75.
Pareidolia thus forms under either loss, and only a patch-level loss keeps it strong
to the last block, as in the four families of \cref{fig:layerwise}.

\section{Further Discussion on Controlled Training}
\label{sec:supp-controlled}

All runs use ViT-S/16 (12 blocks, $d=384$, 6 heads),
ImageNet-100~\cite{deng2009imagenet,tian2020contrastive} (126,689 training images)
and the iBOT code with bf16 autocast.
The two runs of each pair share the data order, augmentation and schedule, and
differ only in the patch-level loss and its masks.
We evaluate the teacher (EMA) model.

\paragraph{From scratch.}
100 epochs, batch 256, peak learning rate $5\times10^{-4}$ with 10 warm-up epochs
and cosine decay, iBOT's official ViT-S/16 recipe (weight decay 0.04 to 0.4,
teacher momentum 0.996 to 1, two $224^2$ global and ten $96^2$ local crops, shared
head with 8192 outputs).
One run sets the patch-loss weight to 0 and uses no masking, which leaves DINO's
image-level loss alone; the other uses weight 1 with iBOT's block masks, which is
iBOT's loss.

\paragraph{From the released iBOT-S.}
20,000 steps of batch 128 (about 20 ImageNet-100 epochs), peak learning rate
$5\times10^{-5}$ with 2,000 warm-up steps and cosine decay, starting from the
official iBOT-S/16 checkpoint with its head and loss centres.
One run keeps only the \cls distillation loss, and the other keeps iBOT's full
loss.

\paragraph{Results.}
\Cref{fig:controlled} shows the pareidolia index and object discovery over training.
From scratch, the gap in the pareidolia index appears by epoch 10 and never
closes, the excess share separates with it (0.21 against 0.06 at the end), and the
run with the patch-level loss is below the other in object discovery at every epoch
from 10 on, while its $k$-NN accuracy on ImageNet-100 is higher (69.7 against 66.3 at
the end).
From the released iBOT-S, the gap in object discovery reaches 9.4 CorLoc by a
quarter of the steps and follows the pareidolia index over training.
At the final checkpoint, the run with the patch-level loss is 5.6 CorLoc lower
from scratch and 8.4 lower from iBOT-S.

\section{Additional Discussion and Analysis}
\label{sec:supp-discussion}

\subsection{What the pareidolic subspace holds}
\label{sec:supp-template}

\begin{figure}[t]
  \centering
  \includegraphics[width=0.9\linewidth]{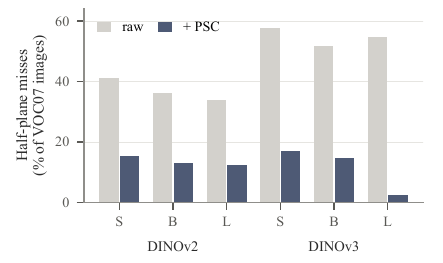}
  \vspace{-12pt}
  \caption{\textbf{Calibration removes most half-plane misses}, the failure in which
  object discovery misses the object with a box along two image borders that covers
  at least 40\% of the image. Share of VOC07 images with such a miss, for the raw
  features (grey) and after \ours (indigo).}
  \label{fig:halfplane}
  \vspace{-12pt}
\end{figure}

\begin{figure}[t]
  \centering
  \includegraphics[width=0.9\linewidth]{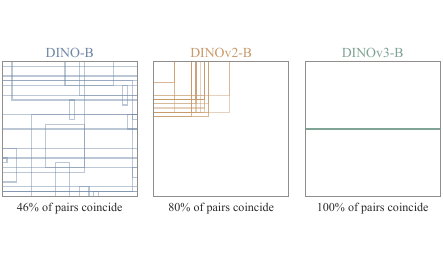}
  \vspace{-12pt}
  \caption{\textbf{On noise, the template decides the cut.} The boxes that TokenCut
  returns on 50 fresh white-noise images, overlaid, for the Base models, and the
  share of box pairs that coincide (IoU of at least 0.5).}
  \label{fig:noiseboxes}
  \vspace{-12pt}
\end{figure}

\begin{figure*}[t]
  \centering
  \includegraphics[width=0.875\textwidth]{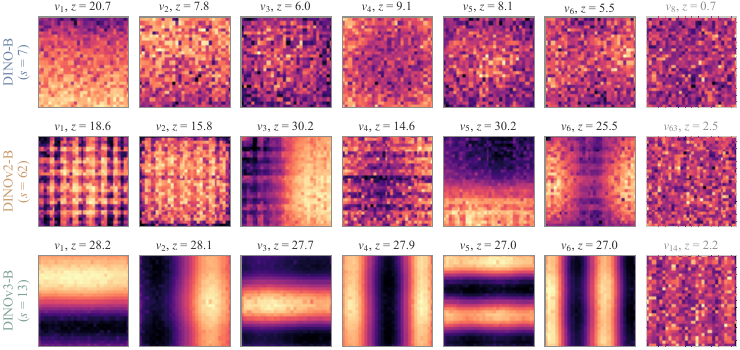}
  \vspace{-6pt}
  \caption{\textbf{The leading directions of the noise response form coherent
  patterns.} Maps of the final-key response of the Base models to one white-noise
  image along its leading directions, with Moran's $z$ (\cref{eq:moranz}) above each
  map. The first six maps of each row are the first six of the $s$ directions of
  the pareidolic subspace, and the last is the first direction after the subspace,
  whose $z$ falls below 3 and ends the leading run.}
  \vspace{-12pt}
  \label{fig:templates}
\end{figure*}

\paragraph{Half-plane misses.}
On natural images, object discovery with the raw features of DINOv2 and DINOv3 often
misses the object with a half-plane box, a box along two image borders that covers at
least 40\% of the image: it does so on 34--58\% of VOC07 images, and after calibration
on 3--17\% (\cref{fig:halfplane}).

\paragraph{The template decides the cut on noise.}
Where the input contains no object, the template decides what a training-free
method returns.
We run TokenCut on 50 fresh white-noise images, each with its own seed, and overlay
the 50 boxes (\cref{fig:noiseboxes}).
On DINO-B, whose template is faint, the boxes scatter over the image, and the pairs
that coincide are mostly whole-image boxes.
On DINOv2-B, TokenCut returns the same top-left box on 46 of the 50 images, and on
DINOv3-B the same half-plane box on all 50.
DINOv2 and DINOv3 thus return a box that does not depend on the input, whereas
the boxes of DINO vary with the noise, in line with the main text: pareidolia is
faint in DINO and strong in DINOv2 and DINOv3 (\cref{sec:share}), and it is in
DINOv2 and DINOv3 that it hinders object emergence (\cref{sec:hinders}).

\paragraph{The template on noise.}
\Cref{fig:templates} shows the maps behind \cref{fig:probe}a one direction at a
time, and the three models differ in how strong a template they impose.
In DINO-B, the directions of the pareidolic subspace are weak gradients amid
noise; in DINOv2-B, they mix grid-aligned patterns with gradients; and in
DINOv3-B, every direction shown is a smooth wave over the image, with Moran's $I$
between 0.96 and 1.00.

\subsection{How the last block carries it}
\label{sec:supp-heads}

\begin{figure}[t]
  \centering
  \includegraphics[width=0.9\linewidth]{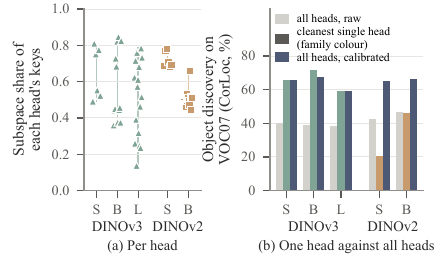}
  \vspace{-12pt}
  \caption{\textbf{In DINOv3 one clean head matches the calibrated full keys; in
  DINOv2 no head does.} (a) Share of the pareidolic subspace in the keys of each
  last-block head (one marker per head). (b) Object discovery on held-out VOC07
  images from all heads of the raw keys, from the single head whose keys hold the
  least of the subspace (chosen without labels), and from all heads after the
  subspace is calibrated.}
  \vspace{-12pt}
  \label{fig:heads}
\end{figure}

\paragraph{Measurement.}
We ask whether the pareidolic subspace sits in every attention head of the last
block or in a few.
For each head, we measure how much of the subspace's key energy it holds, pick the
head that holds the least, a choice that needs no labels, and run object discovery
on that head's keys alone, against the raw full keys and the full keys after
calibration.

\paragraph{Results.}
The two families differ (\cref{fig:heads}a).
In DINOv3-B the heads split: five heads hold 68--85\% subspace in their keys and
seven hold 36--46\%, so half of the heads carry 82\% of the subspace's key energy.
In DINOv2-B every head holds 44--66\%, and half of the heads carry 67\%.
The split decides what a single head can do (\cref{fig:heads}b).
In DINOv3 the cleanest head alone matches the calibrated full keys for all three
sizes (65.8, 71.5 and 59.0 against 65.5, 67.8 and 59.0 CorLoc on held-out VOC07
images) and raises the raw full keys by 21--33 points.
In DINOv2 the cleanest head is no better than the raw full keys (20.2 and 46.2
against 42.8 and 46.8) and falls 20--45 points short of calibration, because no
head is clean.

\paragraph{Reading.}
In DINOv3, the pareidolic subspace is concentrated in a few heads, so the cleanest
head alone recovers object discovery; in DINOv2, it is spread over every head, so
no single head is free of it, while calibrating the subspace recovers object
discovery in both families.

\end{document}